\documentclass{article}

\PassOptionsToPackage{numbers, compress}{natbib}
\usepackage[preprint]{neurips_2026}

\usepackage[utf8]{inputenc} 
\usepackage[T1]{fontenc}    
\usepackage{hyperref}       
\usepackage{url}            
\usepackage{booktabs}       
\usepackage{amsfonts}       
\usepackage{nicefrac}       
\usepackage{microtype}      
\usepackage{xcolor}         

\usepackage[table]{xcolor}
\usepackage{amsmath}
\usepackage{graphicx}
\usepackage{enumitem}
\usepackage{multirow}
\usepackage{multicol}
\usepackage{rotating}
\usepackage{wrapfig}
\usepackage{caption}

\newcommand{\best}[1]{\cellcolor[RGB]{194,241,200}{#1}}
\newcommand{\second}[1]{\cellcolor{orange!20}{#1}}

\title{DegBins: Degradation-Driven Binning for \\ Depth Super-Resolution}

\author{%
Zhiqiang Yan$^{1}$ \quad Zhengxue Wang$^{2}$ \quad Jian Yang$^{2}$ \quad Gim Hee Lee$^{1}$\\
$^1$Department of Computer Science, National University of Singapore\\
$^2$Nanjing University of Science and Technology
}

\begin{document}

\maketitle

\begin{abstract}
Depth super-resolution (DSR) aims to recover a high-resolution (HR) depth map from its low-resolution (LR) counterpart. 
With color image guidance, this task is typically formulated as learning the residual between HR and LR in a low-dimensional feature space. 
However, this additive formulation is insufficient to accurately capture the complex relationship between HR and LR, especially under spatially varying degradations. 
In this paper, we introduce \textbf{DegBins}, a novel DSR framework that leverages degradation-driven binning to adaptively enhance residual modeling. 
Specifically, DegBins reformulates the regression-based DSR as a hybrid classification-regression problem, where the residual depth is represented as a linear combination of discrete depth bins weighted by their learned probability distribution, yielding more flexible and expressive representations. 
Furthermore, DegBins models the degradation relationship between HR and LR in a high-dimensional feature space, enabling adaptive bin range adjustment and probability optimization conditioned on local degradation characteristics. 
To progressively improve reconstruction quality, DegBins adopts a multi-stage refinement scheme, where each stage performs finer-grained bin partitioning and probability updating based on the former estimation. 
This coarse-to-fine design facilitates more accurate depth recovery, particularly in regions with severe degradations or complex structural variations. 
Extensive experiments across five benchmarks demonstrate that DegBins consistently outperforms existing state-of-the-art methods in terms of accuracy, robustness, and generalization. 
\end{abstract}

\section{Introduction}
Depth super-resolution (DSR) is a fundamental task in geometric computer vision, with broad applications in 3D reconstruction~\cite{he2021towards,yan2022learning,de2022learning,zhong2023deep}, augmented reality~\cite{su2019pixel,wang2023g2,yuan2023structure,zhong2021high}, and robotics~\cite{sun2021learning,zhao2022discrete,metzger2023guided}, where precise high-quality depth perception is critical for reliable scene understanding and downstream geometry-aware decision making. 
With guidance from color images, this task is commonly formulated as residual learning~\cite{zhong2023guided,he2021towards,kim2021deformable} between the target high-resolution (HR) depth and the observed low-resolution (LR) depth. 
However, the degradation relationship~\cite{wang2025dornet,yan2024completion} between HR and LR is inherently complex, especially in real-world scenarios where spatially varying blur, noise, sensor imperfections, and misalignment artifacts are prevalent, rendering the naive additive residual formulation insufficient to accurately characterize such a relationship.

\begin{wrapfigure}{r}{0.5\textwidth}
 \centering
 \includegraphics[width=0.5\textwidth]{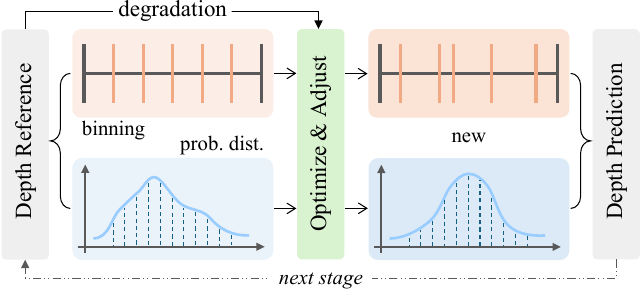}
 \caption{Conceptual core idea of our DegBins.}\label{fig_fig1}
\end{wrapfigure}

To address these limitations, we propose DegBins, which leverages a degradation-driven binning strategy to guide the residual modeling from a distributional perspective instead of a deterministic regression paradigm. 
First, DegBins reformulates the pixel-wise regression-based DSR into a hybrid classification-regression problem. As illustrated in Fig.~\ref{fig_fig1}, the numerical range of residual depth is discretized into multiple depth intervals termed bins. 
The model then predicts a probability distribution over these bins, and computes the refined residual depth as a weighted linear combination 
of the bin centers using the predicted probabilities. 
Compared with conventional DSR approaches, this design enables more flexible and expressive representations of the residual depth. 
Second, in contrast to learning the residual relationship in a limited low-dimensional space, DegBins exploits degradation representations between HR and LR in a high-dimensional space to enable adaptive adjustment of bin ranges and optimization of probability distributions conditioned on local degradation characteristics. This facilitates a more accurate characterization of the relationship by leveraging high-dimensional spatial dependencies to guide those in the low-dimensional space. 
Third, DegBins employs a multi-stage refinement scheme, in which each stage adopts finer-grained bin partitioning and updates the probability distribution based on the former prediction. This coarse-to-fine strategy progressively enhances depth quality and is particularly effective in regions with severe degradation or complex structural variations. 

In summary, our main contributions are as follows:
\begin{itemize}[leftmargin=2em,itemsep=2pt]
    \item 
    We present a degradation-driven binning based framework termed DegBins, which leverages high-dimensional degradation representations to enable adaptive bin range adjustment and probability distribution optimization.
    \item 
    We propose a multi-stage refinement scheme through coarse-to-fine bin partitioning and probability updating based on the former estimation for high-quality depth recovery, especially in heavily degraded or structurally complex regions.
    \item 
    Comprehensive evaluations demonstrate that DegBins achieves superior performances over current leading methods with notable gains in accuracy while maintaining strong robustness and generalization across diverse scenarios.
\end{itemize}

\section{Related Work}
\textbf{Depth Super-Resolution.} 
Early approaches mainly comprise optimization-based~\cite{ferstl2013image} and filtering-based paradigms~\cite{lo2017edge}. They typically rely on hand-crafted priors to restore HR depth, making it difficult to accurately recover geometric boundaries in structurally complex regions. To overcome these limitations, numerous deep learning-based methods~\cite{zhao2023spherical, wang2026spatiotemporal, kang2025c2pd, yan2022learning} are proposed that focus on exploring cross-modal feature fusion between RGB and depth. Such methods~\cite{zhong2021high, wang2024scene, sun2021learning, tang2021bridgenet, wang2023g2} substantially enhance the LR depth by adaptively incorporating geometry-related structural cues from RGB while suppressing irrelevant texture copying.
For example, \citet{zhao2022discrete} introduce discrete cosine transform to transfer modality-shared features from RGB to depth. SGNet~\cite{wang2024sgnet} presents a structure-guided framework to aggregate depth-consistent high-frequency components across spatial, gradient, and frequency domains. Additionally,~\citet{zheng2025decoupling} design a geometry-decoupled network to disentangle fine-grained and global geometric knowledge. ~\citet{zhong2025dual} develop a cross-modality neural architecture search framework to automatically design efficient fusion models. 

\noindent\textbf{Degradation Modeling.} 
Recent progress in color image restoration has been largely driven by more expressive degradation modeling strategies \cite{yin2022conditional,zhang2021designing,wang2021unsupervised}. For example, \cite{zhang2021designing} design a realistic and practically deployable degradation pipeline to mimic complex real-world corruptions in blind super-resolution. In a similar vein, \cite{zhou2023learning} introduce a degradation-adaptive regression framework that predicts region-specific degradation for local patches. 
\cite{liang2022efficient} propose a model in which network parameters are dynamically modulated according to the estimated degradation characteristics. 
Beyond single-degradation settings, a number of approaches \cite{zhang2023all,li2022all,zhang2023ingredient} aim to handle multiple degradation types within a unified framework. For example, \citet{zhang2023all} present a multi-degradation restoration network that progressively captures degradation patterns through clustering, eliminating the need for explicit degradation priors. Similarly, \citet{zhang2023ingredient} formulate an ingredient-oriented representation to explicitly characterize inter-image degradation relationships. More recently, \citet{yao2024neural} propose a neural framework that captures latent degradation factors and disentangles mixed degradations for improved restoration. 
DORNet~\cite{wang2025dornet} adaptively addresses unknown degradations in real-world scenes using implicit degradation representations. 

\noindent\textbf{Bin-Based Depth Learning.} 
Monocular depth estimation is typically formulated as a regression problem~\cite{eigen2014depth,shao2023nddepth,wang2024dcdepth}, where pixel-wise depth values are directly predicted under a regression loss. DORN~\cite{fu2018deep} first reformulates this task as a per-pixel classification problem by discretizing the continuous depth range into a set of intervals (bins). Based on this idea, AdaBins~\cite{bhat2021adabins} adopts a hybrid classification–regression paradigm to predict the probability distribution over globally shared bins and compute the final depth via a linear combination. BinsFormer~\cite{li2024binsformer} further introduces a sufficient interaction between probability distribution and bin predictions, and LocalBins~\cite{bhat2022localbins} replaces globally shared bins with pixel-wise adaptive bins. Moreover, IEBins~\cite{shao2023iebins} introduces elastic bins which progressively refine the bin ranges in a multi-stage manner. 
During the process of high-resolution depth reconstruction, all these bin-based depth estimation approaches offer valuable insights, particularly when degradation representations are available.

\begin{figure}[t]
\centering
\includegraphics[width=0.99\columnwidth]{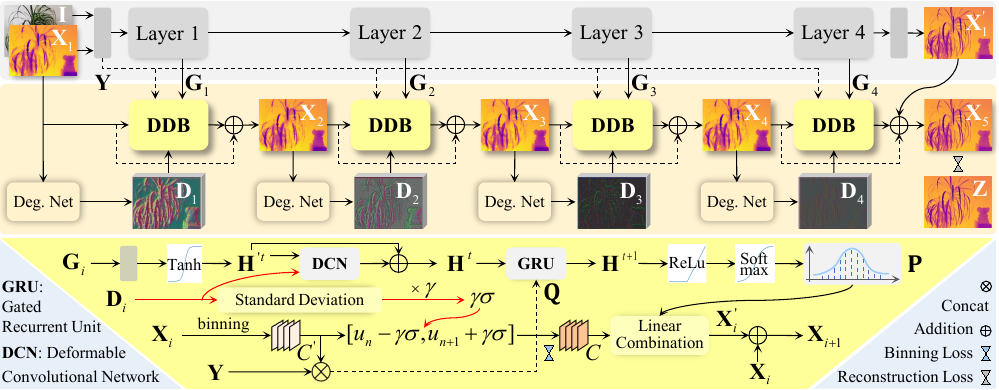}\\
\caption{Framework of DegBins, where DDB represents the degradation-driven binning strategy.
}\label{fig_pipeline}
\end{figure}

\section{Methodology}
\subsection{Overview}
Inspired by prior works~\cite{fu2018deep,bhat2021adabins}, we reformulate the pixel-wise regression-based DSR as a hybrid classification–regression paradigm, enabling more flexible and expressive representations that improve the model’s generalization capability. 
Fig.~\ref{fig_pipeline} illustrates the pipeline of DegBins, which primarily comprises two components: the gray part representing the basic feature encoding layers and the orange part corresponding to the progressive depth prediction blocks. 
In the gray part, we employ the pre-trained DuCos~\cite{yan2025ducos} as the backbone, which takes a color image $\mathbf{I}$ and a LR depth map $\mathbf{X}_1$ as input, and outputs a fused context feature $\mathbf{Y}$, layer-specific features $\{\mathbf{G}_1, \mathbf{G}_2, \mathbf{G}_3, \mathbf{G}_4\}$, and an initial depth estimate $\mathbf{X}_{1}^{'}$. 
In the orange part, the LR depth $\mathbf{X}_1$ is first discretized into multiple bins~\cite{bhat2021adabins,bhat2022localbins,shao2023iebins}. The network then infers high-dimensional degradation representations $\{\mathbf{D}_1, \mathbf{D}_2, \mathbf{D}_3, \mathbf{D}_4\}$ via pre-trained degradation blocks~\cite{wang2025dornet}. Conditioned on these degradation representations, the proposed degradation-driven binning (DDB) strategy adaptively adjusts the bin ranges and their associated probability distributions, yielding finer-grained depth outputs. This process is iteratively performed over four stages, where the depth output at each stage serves as the input to the next, ultimately producing the final depth prediction $\mathbf{X}_5$. Subsequently, Sec.~\ref{sec_ddb} introduces the DDB strategy together with its multi-stage refinement scheme. Sec.~\ref{sec_loss} then introduces the overall loss functions.

\subsection{Degradation-Driven Binning}
\label{sec_ddb}
As shown in Fig.~\ref{fig_pipeline}, in addition to the LR depth $\mathbf{X}_1$, the context feature $\mathbf{Y}$ and the layer feature $\mathbf{G}_i$ extracted by the pretrained backbone $\mathcal{F}(\cdot)$~\cite{yan2025ducos}, we further estimate the degradation relationship~\footnote{Based on the assumption that a degradation relationship exists between the low-quality depth $\mathbf{X}$ and target depth $\mathbf{\hat{X}}$, defined as $\mathbf{X} = \mathbf{K}\mathbf{\hat{X}} + \mathbf{n}$, where $\mathbf{K}$ is a degradation kernel derived from the degradation representation $\mathbf{D}$ and $\mathbf{n}$ denotes noise, DORNet models $\mathbf{D}$ in a self-supervised manner by minimizing the discrepancy between $\mathbf{X}$ and $\mathbf{K}\mathbf{\hat{X}}$.} 
between the coarse depth $\mathbf{X}_i$ and the final prediction $\mathbf{X}_5$ using the pretrained degradation learning block $\mathcal{D}(\cdot)$ from DORNet~\cite{wang2025dornet}. 
This process is formulated as:
\begin{subequations}
\begin{align}
&\mathbf{Y}, \mathbf{G}_{i}=\mathcal{F} (\mathbf{I},\mathbf{X}_1), \\
&\mathbf{D}_{i}=\mathcal{D}(\mathbf{X}_i,\mathbf{X}_5), \quad \text{for} \quad i=1,2,3,4.
\end{align}
\end{subequations}
Given these three inputs, the subsequent description of DDB is organized into three aspects: \textbf{(1)} decomposing depth into bin representations, \textbf{(2)} adaptively adjusting bin ranges and optimizing the probability distribution under degradation guidance, and \textbf{(3)} leveraging the former depth estimation to progressively update the subsequent prediction. 

\textbf{(1)} In the $i$-th stage, DDB first discretizes the continuous coarse depth $\mathbf{X}_i$ into multiple intervals~\cite{bhat2021adabins,bhat2022localbins,shao2023iebins}. Let $v_{min}(p)$ and $v_{max}(p)$ denote the minimum and maximum of $\mathbf{X}_i$ at pixel $p$, respectively. Then, the pixel-wise depth range $[v_{min}(p),v_{max}(p)]$ is uniformly divided into $N$ bins, yielding the width $W(p)$ of each bin and the upper bound $u_n(p)$ of the $n$-th bin:
\begin{subequations}
\begin{align}
&W(p)=\frac{v_{max}(p)-v_{min}(p)}{N}, \\
&u_n(p)=v_{min}(p) + nW(p),  \quad \text{for} \quad n=0,1,\cdots,N;
\end{align}
\end{subequations}
where the minimum and maximum values adaptively vary with pixels of the coarse depth, and $N$ is set to 32 unless otherwise specified. 
Since discrete bins cannot be directly used for depth prediction, their center values are adopted as the corresponding depth candidates:
\begin{equation}
C_n^{'}(p)=\frac{u_n(p) + u_{n+1}(p)}{2}, \quad \text{for} \quad n=0,1,\cdots,N-1.
\end{equation}
Based on these bin centers, weighted by their learned probability distribution $\mathbf{P}_n^{'}$ (updated by Eq.~\eqref{eq_prob}), the depth at the pixel $p$ in $\mathbf{X}_{i}^{'}$ is computed as:
\begin{equation}\label{eq_linear_comb}
\mathbf{X}_{i}^{'}(p)=\sum_{i=0}^{N-1} C_n^{'}(p) \cdot \mathbf{P}_n^{'}(p),
\end{equation}
where $\mathbf{P}_n^{'}$ denotes the $n$-th depth probability.

\textbf{(2)} \textbf{Bin Range Adjustment.}  
The target bin is defined as the interval whose boundaries enclose the predicted depth value, \textit{i.e.}, the bin for which $\mathbf{X}_{i}^{'}(p)$ falls within its edges: 
\begin{equation}
u_n(p) \leq \mathbf{X}_{i}^{'}(p) \leq u_{n+1}(p)
\end{equation}
which is denoted as $\mathbf{X}_{i}^{'}(p) \in [u_n(p), u_{n+1}(p)]$. As illustrated in the yellow region of Fig.~\ref{fig_pipeline}, we leverage the degradation representation $\mathbf{D}_i$, which characterizes the likelihood of depth inaccuracy, to adaptively adjust this interval. Motivated by~\cite{hu2022uncertainty,yan2023distortion,shao2023iebins}, we calculate the variance of the degradation:
\begin{equation}
V(p) = \frac{1}{|\mathcal{N}(p)|} \sum_{q \in \mathcal{N}(p)} \left(\mathbf{D}(q) - \mu(p)\right)^2,
\end{equation}
where $\mathcal{N}(p)$ denotes a $k \times k$ local neighborhood (with $k=3$ by default) centered at pixel $p$, and $\mu(p)=\frac{1}{|\mathcal{N}(p)|} \sum_{q \in \mathcal{N}(p)} \mathbf{D}(q)$ denotes the corresponding local mean. The standard deviation is then computed as $\sigma(p)=\sqrt{V(p)}$. Given the target bin range $[u_n(p), u_{n+1}(p)]$ and the standard deviation $\sigma(p)$, we can acquire the new range:
\begin{equation}\label{eq_new_range}
[u_n(p)-\gamma \sigma (p),u_{n+1}(p)+\gamma \sigma (p)],
\end{equation}
where $\gamma$ is a coefficient controlling the error tolerance and is set to 0.2. Thus we can renew $v_{min}(p)$ and $v_{max}(p)$ as $u_n(p)-0.2 \sigma (p)$ and $u_{n+1}(p)+0.2 \sigma (p)$, yielding updated depth candidate $C_n$ and prediction via Eq.~\eqref{eq_linear_comb}. 
For pixels with large discrepancies from the ground-truth (GT) depth, the standard deviation increases accordingly, yielding a more tolerant target bin with higher immunity that helps mitigate errors. 

\begin{figure}[t]
 \centering
 \includegraphics[width=0.91\columnwidth]{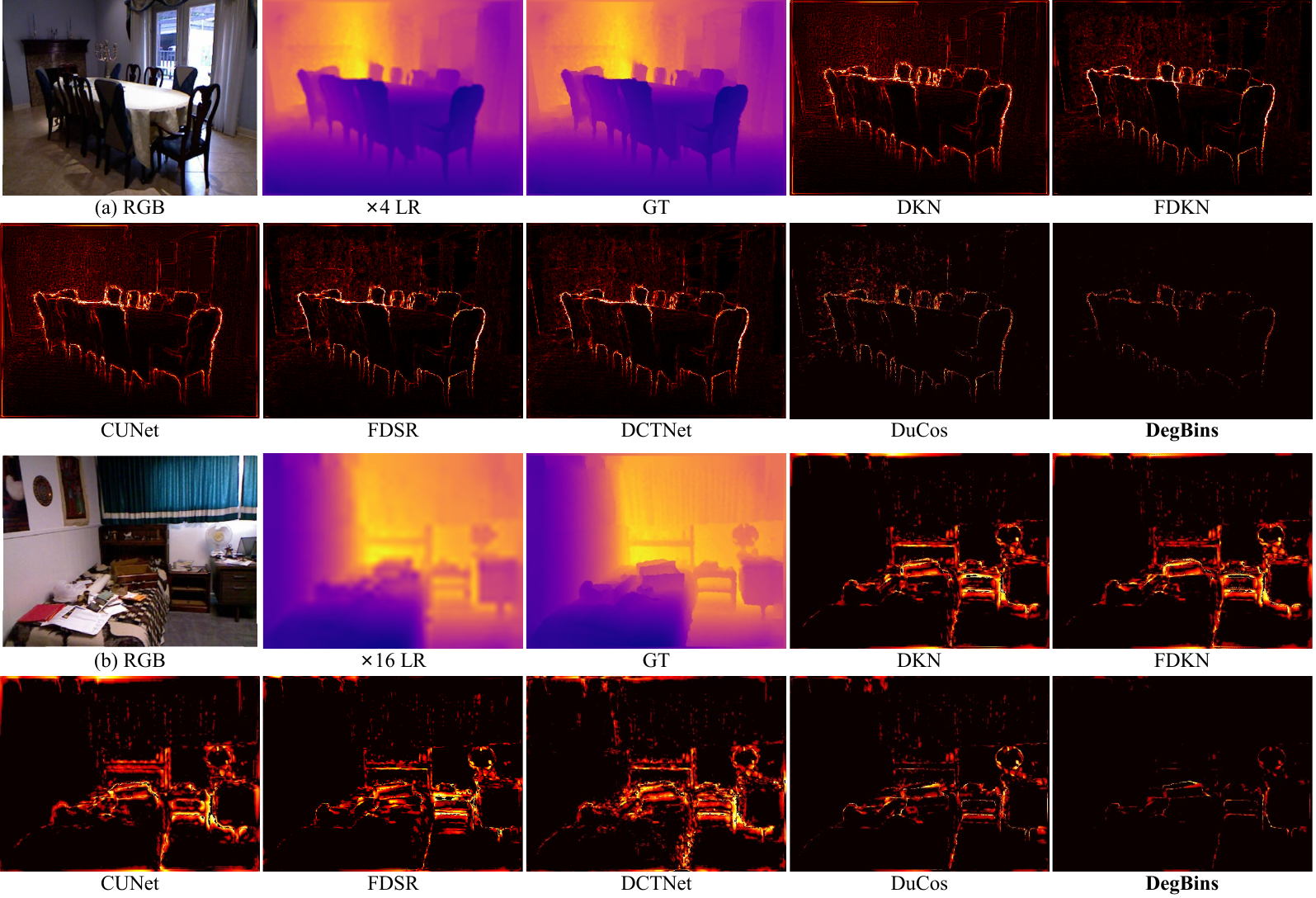}\\
 \vspace{-2pt}
 \caption{Error map comparisons on synthetic NYU v2 with $\times 4$ and $\times 16$ upscaling factors.
 }\label{fig_vis_syn_nyu}
 \vspace{-3pt}
\end{figure}

\textbf{Probability Optimization.} 
Unlike prior methods~\cite{bhat2021adabins,bhat2022localbins,li2024binsformer} that rely on heavy transformer modules, we efficiently estimate the probability distribution using a gated recurrent unit~\cite{teed2020raft,shao2023iebins} (GRU), which propagates contextual information from previous states. 
Specifically, we first project the depth candidates $C$ into a feature space using four 
$3\times 3$ convolutions, each followed by a ReLU activation. The resulting feature is then concatenated with the context feature $\mathbf{Y}$ to form one of the GRU input $\mathbf{Q}$. 
Another key input, namely the hidden state $\mathbf{H}^{'t}$, is generated through a $3\times 3$ convolution followed by a $\tanh$ activation. We further incorporate the degradation priority into this hidden representation using a deformable convolutional network~\cite{zhu2019deformable} (DCN) $\mathcal{J}(\cdot)$, where both the offsets and modulation scalars are derived from the degradation $\mathbf{D}_i$, yielding a new hidden state:
\begin{equation}
\mathbf{H}^{t}=\mathbf{H}^{'t}+\mathcal{J}(\mathbf{H}^{'t}, \mathbf{D}_i).
\end{equation}
Given the input $\mathbf{Q}$ and the updated hidden state $\mathbf{H}^{t}$, the hidden state at the next time step is computed as $\mathbf{H}^{t+1}=\mathcal{G}(\mathbf{Q}, \mathbf{H}^{t})$. Then we obtain the probability via:
\begin{equation}\label{eq_prob}
\mathbf{P}=\mathcal{F}_{r}(\mathcal{F}_{t}(\mathbf{H}^{t+1})),
\end{equation}
where $\mathcal{F}_{r}(\cdot)$ and $\mathcal{F}_{t}(\cdot)$ denote the ReLU and softmax activation functions, respectively. Consequently, by incorporating Eq.~\eqref{eq_new_range} and Eq.~\eqref{eq_prob}, Eq.~\eqref{eq_linear_comb} can be reformulated as:
\begin{equation}\label{eq_linear_comb_updated}
\mathbf{X}_{i}^{'}(p)=\sum_{i=0}^{N-1} C_n(p) \cdot \mathbf{P}_n(p).
\end{equation}

\textbf{(3)} We summarize all the preceding modeling in this section as a unified DDB process, where the inputs are the coarse depth $\mathbf{X}_i$, context feature $\mathbf{Y}$, layer feature $\mathbf{G}_i$, and degradation $\mathbf{D}_i$, and the output is $\mathbf{X}_{i+1}$. Denoting the DDB process function as $\mathcal{P}(\cdot)$, the formulation for the progressive scheme is given as follows:
\begin{equation}\label{eq_ddb}
\mathbf{X}_{i+1}=\mathbf{X}_i + \mathcal{P}(\mathbf{X}_i,\mathbf{G}_i,\mathbf{D}_i,\mathbf{Y}).
\end{equation}
The depth is gradually refined based on its previous estimation. In short, we aim to mitigate the limitations of residual-based DSR models in capturing the complex relationship between HR and LR by leveraging a DDB strategy built upon high-dimensional degradation representations.

\subsection{Loss Function}
\label{sec_loss}
Given the set of bin centers as $\mathcal{C}$, HR depth $\mathbf{X}$, and GT depth $\mathbf{Z}$ with $m$ valid pixels, the reconstruction loss function is written as $\mathcal{L}_{rec} = \frac{1}{m} \| \mathbf{X} - \mathbf{Z} \|_1$. Following~\cite{bhat2021adabins,bhat2022localbins}, we use the bi-directional Chamfer loss~\cite{fan2017point} as a binning regularizer: $\mathcal{L}_{bin}=\sum_{z \in \mathbf{Z}} \min _{c\in \mathcal{C}} \| c - z \|^2 + \sum_{c \in \mathcal{C}} \min _{z\in \mathbf{Z}} \| c - z \|^2$. 
Consequently, the total loss function is defined as:
\begin{equation}
\mathcal{L}=\mathcal{L}_{rec} + \alpha \mathcal{L}_{bin},
\end{equation}
where $\alpha$ denotes a balance coefficient that is empirically set to 0.1.

\begin{table}[t]
\caption{Results on synthetic DSR benchmarks. The \colorbox[RGB]{194,241,200}{best} and \colorbox{orange!20}{second-best} metrics are highlighted.
}\label{tab_syn}
\centering
\Large
\renewcommand\arraystretch{0.7}
\resizebox{1\linewidth}{!}{
\begin{tabular}{l|c|ccccccccccccccc}
\toprule 
\multirow{2}{*}{Method} &\multirow{2}{*}{Scale}   &\multicolumn{3}{c}{Middlebury}   &\multicolumn{3}{c}{Lu}  &\multicolumn{3}{c}{NYU v2}   &\multicolumn{3}{c}{RGB-D-D}  &\multicolumn{3}{c}{TOFDSR} \\
\cmidrule(lr){3-5}\cmidrule(lr){6-8}\cmidrule(lr){9-11}\cmidrule(lr){12-14} \cmidrule(lr){15-17} 
 & &RMSE &MAE &$\delta _{1}  $     &RMSE &MAE &$\delta _{1}  $  
 &RMSE &MAE &$\delta _{1}  $     &RMSE &MAE &$\delta _{1}  $ &RMSE &MAE &$\delta _{1}  $\\ \midrule

DJF           &  &1.14 &0.63 &98.56       &1.08 &0.44 &99.00               &2.32 &0.83 &99.49      &1.18 &0.36 &99.64   &2.99 &0.66 &98.74   \\
DJFR         & \multirow{10}{*}{$\times 2$}                            &1.22 &0.58 &98.42       &1.39 &0.45 &98.94               &1.87 &0.51 &99.68      &0.98 &0.27 &99.74   &1.58 &0.28 &99.57   \\
CUNet       &                             &1.01 &0.58 &98.78       &1.06 &0.50 &99.10               &1.50 &0.46 &99.81      &0.84 &0.25 &99.81   &1.69 &0.33 &99.49   \\
FDKN   &                             &1.35 &0.62 &98.39       &1.59 &0.49 &98.91               &2.05 &0.50 &99.63      &1.04 &0.26 &99.70   &1.85 &0.36 &99.32   \\
DKN    &                             &1.32 &0.62 &98.48       &1.48 &0.48 &99.00               &1.95 &0.49 &99.66      &0.98 &0.24 &99.74   &1.87 &0.36 &99.33   \\
FDSR       &                             &0.94 &0.51 &98.89       &0.94 &0.39 &99.14               &1.74 &0.57 &99.70      &0.91 &0.26 &99.76   &1.55 &0.40 &99.24 \\
DCTNet  &                             &1.01 &0.53 &98.90      &1.03 &0.59 &99.36              &1.59 &0.59 &99.76      &0.89 &0.28 &99.79   &0.80 &0.23 &99.87   \\
DADA   &                             &1.28 &0.60 &98.47       &1.45 &0.48 &99.07               &2.00 &0.54 &99.65      &1.04 &0.27 &99.71   &1.73 &0.37 &99.26   \\
DuCos                  &              &\second{0.81} &\second{0.46} &\second{99.28}       &\second{0.65} &\second{0.26} &\second{99.68}           &\second{1.20} &\second{0.37} &\second{99.88}      &\second{0.74} &\second{0.22} &\second{99.85}   &\second{0.52} &\second{0.13} &\second{99.94} \\ 
\textbf{DegBins} & & \best{0.76} & \best{0.43} & \best{99.51} & \best{0.57} & \best{0.22} & \best{99.74} & \best{1.12} & \best{0.30} & \best{99.94} & \best{0.68} & \best{0.19} & \best{99.90} & \best{0.46} & \best{0.09} & \best{99.98} \\

\midrule

DJF           & &1.93 &1.11 &97.11                               &1.93 &1.11 &98.20              &3.60 &1.67 &99.05                  &1.63 &0.74 &99.39       &3.75 &1.14 &98.18       \\
DJFR         & \multirow{11}{*}{$\times 4$}                            &1.83 &1.04 &96.91                               &1.85 &1.02 &98.36              &3.27 &1.14 &99.04                  &1.54 &0.48 &99.35       &2.97 &0.58 &99.00       \\
CUNet       &                             &1.61 &0.98 &97.79                   &1.73 &0.97 &98.48              &3.22 &1.44 &99.21                  &1.52 &0.65 &99.43       &3.57 &1.13 &98.19       \\
FDKN   &                             &1.80 &0.80 &97.69                               &2.11 &0.58 &98.53              &3.28 &0.93 &99.28                  &1.56 &0.42 &99.42       &2.80 &0.58 &98.96       \\
DKN    &                             &1.77 &0.78 &97.77                               &2.05 &0.56 &98.58  &3.15 &0.90 &99.34                  &1.50 &0.41 &99.49       &2.73 &0.56 &99.04    \\
FDSR       &                             &1.72 &0.84 &97.69                               &1.95 &0.56 &98.52  &2.94 &0.85 &99.39                  &1.41 &0.40 &99.51       &2.41 &0.51 &99.07    \\
DCTNet  &                             &1.66 &0.77 &97.86       &1.85 &0.57 &98.56              &2.90 &0.99 &99.37                  &1.49 &0.44 &99.45       &2.86 &0.59 &98.98    \\
DADA   &                             &1.82 &0.83 &97.45                               &2.12 &0.66 &98.35              &3.08 &0.96 &99.33                  &1.63 &0.46 &99.39       &2.85 &0.68 &98.57      \\
DORNet   &                              &1.69 &0.80 &97.76                   &1.78 &0.56 &98.70  &2.79 &0.85 &99.44      &1.45 &0.41 &99.50    &2.28 &0.36 &99.49  \\ 
DuCos  &  &\second{1.45} &\second{0.68} &\second{98.30}  &\second{1.38}  &\second{0.41}  &\second{99.08}  &\second{2.60} &\second{0.81} &\second{99.49}  &\second{1.27} &\second{0.36} &\second{99.60}   &\second{2.09} &\second{0.33} &\second{99.58} \\ 
\textbf{DegBins} & & \best{1.36} & \best{0.64} & \best{98.42} & \best{1.33} & \best{0.39} & \best{99.14} & \best{2.57} & \best{0.78} & \best{99.51} & \best{1.25} & \best{0.32} & \best{99.65} & \best{1.97} & \best{0.30} & \best{99.64} \\
\midrule

DJF           & &3.09 &1.46 &94.10                          &3.58 &1.22 &95.60            &5.56 &2.30 &97.70                  &2.58 &0.96 &98.29           &5.59 &1.71 &96.29  \\
DJFR         & \multirow{11}{*}{$\times 8$}                            &2.82 &1.25 &95.31                          &3.24 &1.07 &95.86            &5.20 &1.94 &98.21                  &2.61 &0.93 &98.25           &5.11 &1.35 &97.16  \\
CUNet       &                             &2.86 &1.46 &94.44                          &2.85 &1.25 &96.32            &5.50 &2.23 &97.78                  &2.35 &0.92 &98.60           &5.14 &1.64 &96.67   \\
FDKN   &  &2.51 &1.13 &96.26   &\second{2.67} &0.85 &97.39  &4.93 &1.67 &98.65    &2.25 &0.70 &98.86        &\second{4.40} &0.96 &98.15      \\
DKN    &                             &2.43 &1.12 &96.26              &2.88 &0.88 &97.13            &4.88 &1.71 &98.62                  &2.33 &0.74 &98.76        &4.54 &1.06 &97.87      \\
FDSR       &                             &2.41 &1.13 &96.24  &2.69 &0.86 &97.35            &4.82 &1.69 &98.57    &2.25 &0.73 &98.72   &\best{4.28} &0.95 &98.06          \\
DCTNet  &                             &2.75 &1.31 &95.25                          &3.07 &1.18 &96.38            &4.90 &2.12 &98.28                  &2.47 &0.92 &98.60      &5.38 &1.57 &97.01       \\
DADA   &                             &2.77 &1.27 &95.01                          &3.76 &1.17 &96.01            &4.83 &1.86 &98.34                  &2.81 &0.93 &98.29      &5.86 &1.64 &96.15     \\
DORNet   &                              &2.34 &1.03 &96.85  &2.81 &0.82 &97.50              &4.76 &1.58 &98.81                   &2.33 &0.69 &98.80    &4.81 &0.97 &98.14  \\ 
DuCos  &  &\second{2.23} &\second{0.97} &\second{97.07}     &\second{2.67} &\second{0.72} &\second{97.86}    &\second{4.61} &\second{1.52} &\second{98.86}    &\second{2.23} &\second{0.66} &\second{98.89}  &4.60 &\second{0.92} &\second{98.30}\\ 
\textbf{DegBins} & & \best{2.20} & \best{0.96} & \best{97.12} & \best{2.63} & \best{0.68} & \best{97.99} & \best{4.57} & \best{1.45} & \best{98.87} & \best{2.14} & \best{0.61} & \best{98.94} & 4.42 & \best{0.86} & \best{98.40} \\

\midrule

DJF           & &5.50 &2.92 &84.66       &6.53 &2.69 &88.91            &9.82 &4.73 &93.05                  &4.46 &2.04 &94.99        &8.19 &3.49 &90.79      \\
DJFR         & \multirow{11}{*}{$\times16$}                            &5.16 &2.61 &86.39       &6.46 &2.38 &89.66            &9.50 &4.28 &93.63                  &4.36 &1.84 &95.48        &8.06 &3.06 &92.20      \\
CUNet       &                             &4.72 &2.40 &88.45       &5.63 &2.17 &91.33            &8.63 &3.88 &94.55                  &3.81 &1.58 &96.36        &7.36 &2.73 &93.40      \\
FDKN   &                             &4.42 &2.10 &90.80       &5.48 &1.91 &92.24            &7.97 &3.43 &95.76                  &3.71 &1.45 &96.81        &7.16 &2.29 &94.78      \\
DKN    &                             &4.17 &1.97 &91.67       &5.44 &1.90 &91.94            &7.70 &3.28 &96.06                  &3.70 &1.39 &97.06        &7.24 &2.20 &95.26      \\
FDSR       &                             &3.97 &1.81 &92.52       &5.23 &1.67 &93.44            &\second{7.29} &2.91 &96.86                  &\second{3.44} &1.24 &97.37        &\second{6.85} &1.91 &95.96     \\
DCTNet  &                             &5.07 &2.59 &86.91       &5.83 &2.25 &90.19            &9.10 &4.24 &94.04                  &4.13 &1.66 &96.30        &8.04 &2.77 &93.45     \\
DADA   &                             &4.11 &2.06 &90.09       &6.19 &2.22 &90.95            &7.99 &3.54 &95.64                  &4.01 &1.59 &96.71        &7.79 &2.70 &93.25     \\
DORNet   &                             &4.32 &1.98 &91.78                            &5.27 &1.75 &93.21               &7.60 &3.00 &96.81                   &3.88 &1.34 &97.23    &7.94 &2.19 &95.46  \\ 
DuCos   &   &\second{3.96} &\second{1.69} &\second{93.52}    &\second{5.18} &\second{1.59} &\second{93.89}    &7.37 &\second{2.82} &\second{97.12}      &\second{3.44} &\second{1.16} &\second{97.61}   &6.90 &\second{1.77} &\second{96.51}\\ 
\textbf{DegBins} & & \best{3.74} & \best{1.60} & \best{94.02} & \best{4.93} & \best{1.48} & \best{94.27} & \best{7.11} & \best{2.65} & \best{97.36} & \best{3.29} & \best{1.10} & \best{97.82} & \best{6.73} & \best{1.68} & \best{96.97} \\

\midrule

DJF           & &9.14 &5.33 &70.35       &11.03 &5.16 &77.59            &16.14 &8.29 &84.68                  &7.29 &3.85 &87.33         &12.29 &6.56 &82.85    \\
DJFR         & \multirow{11}{*}{$\times32$}                            &8.26 &4.79 &72.45       &9.98  &4.62 &78.98            &15.22 &7.76 &85.77                  &6.60 &3.55 &89.20         &11.48 &5.76 &85.03     \\
CUNet       &                             &8.48 &4.87 &72.75       &10.56 &4.90 &78.61            &15.42 &7.84 &85.64                  &6.90 &3.56 &89.13         &11.81 &6.01 &84.69     \\
FDKN   &                             &7.73 &4.40 &75.11       &9.24  &4.37 &79.71            &14.06 &7.20 &87.17                  &6.30 &3.24 &90.76         &10.74 &5.13 &86.99     \\
DKN    &                             &7.35 &4.14 &76.97       &9.50  &4.59 &78.94            &13.96 &7.10 &87.47                  &6.12 &3.05 &91.29         &10.46 &4.80 &87.47     \\
FDSR       &                             &7.79 &4.31 &76.11       &\second{9.15}  &\second{4.21} &\second{80.35}            &13.55 &6.96 &87.71                  &6.05 &3.00 &91.32         &9.64 &4.38 &88.40    \\
DCTNet  &                             &9.02 &5.22 &70.59       &10.45 &5.18 &76.27            &15.99 &8.51 &83.93                  &6.45 &3.42 &89.72         &10.87 &5.39 &85.81    \\
DADA   &                             &\best{6.84} &\second{3.75} &\second{78.99}       &9.27 &4.44 &79.43             &12.94 &6.74 &88.40                  &5.52 &2.84 &92.37          &9.60 &4.49 &88.22   \\     
DORNet   &                              &7.56 &4.25 &76.98   &\second{9.15} &4.42 &79.65               &13.81 &6.92 &88.25                   &5.77 &2.72 &92.50    &10.14 &4.14 &89.62  \\ 
DuCos   &                                 &7.69 &4.14 &78.63   &10.40 &4.68 &78.56    &\second{12.27} &\second{5.60} &\second{92.00}   &\second{4.90} &\second{2.03} &\second{95.40}     &\second{9.22} &\second{3.03} &\second{93.30} \\ 
\textbf{DegBins} & & \second{6.87} & \best{3.70} & \best{81.15} & \best{9.12} & \best{3.93} & \best{81.61} & \best{12.06} & \best{5.47} & \best{92.34} & \best{4.84} & \best{1.99} & \best{95.50} & \best{9.16} & \best{2.98} & \best{93.37} \\
\bottomrule
\end{tabular}}
\vspace{-5pt}
\end{table}

\section{Experiment}
\noindent \textbf{Dataset.} 
Following DuCos~\cite{yan2025ducos}, all methods are trained on the fully simulated Hypersim~\cite{roberts2021hypersim} dataset and evaluated on test splits of multiple unseen DSR benchmarks, including TOFDSR~\cite{yan2024tri, yan2025tri}, RGB-D-D~\cite{he2021towards}, NYUv2~\cite{silberman2012indoor}, Middlebury~\cite{hirschmuller2007evaluation,scharstein2007learning}, and Lu~\cite{lu2014depth}. Consistent with prior works~\cite{he2021towards,zhao2023spherical,yan2025ducos}, we construct LR depth inputs via bicubic downsampling of GT depth annotations across these datasets. In addition, TOFDSR and RGB-D-D further provide real-world LR depth measurements, allowing for a more comprehensive evaluation of model generalization under real sensing conditions. Refer to our Appendix~\ref{appendix} for metrics and implementation details.

\subsection{Comparison with State-of-the-Arts}
In this section, we compare DegBins with various methods: DJF~\cite{li2016deep}, DJFR~\cite{li2019joint}, CUNet~\cite{deng2020deep}, DKN~\cite{kim2021deformable}, FDKN~\cite{kim2021deformable}, FDSR~\cite{he2021towards}, DCTNet~\cite{zhao2022discrete}, SUFT~\cite{shi2022symmetric}, SFG~\cite{yuan2023structure}, DADA~\cite{metzger2023guided}, DORNet~\cite{wang2025dornet}, and DuCos~\cite{yan2025ducos}.
We further evaluate their generalization performance across five settings: synthetic DSR, arbitrary-scale DSR, cross-scale DSR, real-world DSR, and compressed DSR.

\begin{figure}[t]
 \centering
\includegraphics[width=0.92\columnwidth]{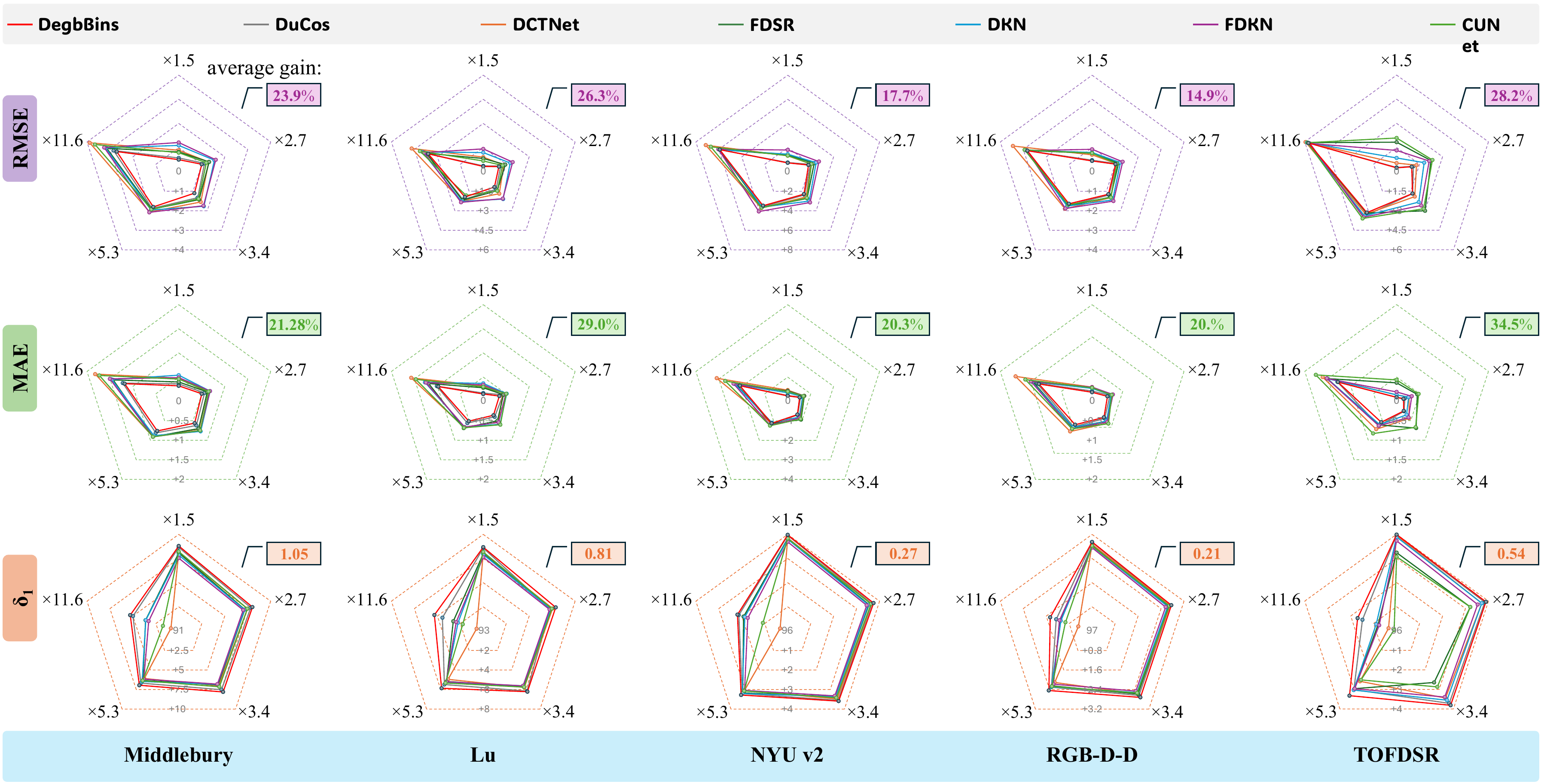}\\
\vspace{-3pt}
 \caption{Quantitative comparisons of arbitrary scaling factors on five synthetic DSR datasets.}\label{fig_arbitrary}
 \vspace{-15pt}
\end{figure}

\noindent\textbf{Synthetic DSR.} 
Tab.~\ref{tab_syn} reports quantitative comparisons across five scaling factors on multiple public synthetic datasets. It is evident that our DegBins consistently achieves superior effectiveness and robustness on all evaluation metrics. For small scaling factors ($\times2$ and $\times4$), DegBins obtains the lowest RMSE and MAE while maintaining $\delta _{1} $ accuracy near or above $99\%$, highlighting its advantage in both detail restoration and pixel-wise precision. For example, compared to the suboptimal method DuCos, our DegBins reduces RMSE by $0.08cm$ and MAE by $0.07cm$ on the $\times2$ NYU v2 dataset. As the scaling factor increases, reconstructing high-quality HR depth becomes more challenging. Nevertheless, our method delivers even larger performance gains, surpassing the second-best approach by $0.18cm$ and $0.21cm$ in RMSE on $\times16$ and $\times32$ NYU v2 dataset, respectively. Furthermore, Fig.~\ref{fig_vis_syn_nyu} presents visual comparisons across different scaling factors, confirming that DegBins successfully recovers accurate depth and with fewer prediction errors.

\noindent\textbf{Arbitrary-scale DSR.} 
In real-world applications, scaling factors are often non-fixed, including both integer and non-integer values. To evaluate the generalization ability of our method under arbitrary 
scales, we further conduct experiments with multiple non-integer factors. As shown in Fig.~\ref{fig_arbitrary}, DegBins consistently  delivers  highly competitive performance across all datasets, metrics, and scales. For instance, compared with all competing methods, DegBins achieves an average improvement of $23.9\%$ in RMSE, $21.28\%$ in MAE, and $1.05$ percent points in $\delta _{1} $ on the Middlebury dataset.

\begin{wraptable}{r}{0.59\textwidth}
\vspace{-15pt}
\caption{Cross-scale DSR on synthetic RGB-D-D and Lu.
}\label{tab_cross_scale_supp}
\vspace{-7pt}
\centering
\renewcommand\arraystretch{0.7}
\resizebox{1\linewidth}{!}{
\begin{tabular}{l|c|cccccc}
\toprule 
\multirow{2}{*}{Method}  & \multirow{2}{*}{scale}   &\multicolumn{3}{c}{RGB-D-D}   &\multicolumn{3}{c}{Lu}  \\
\cmidrule(lr){3-5}\cmidrule(lr){6-8}
&  & RMSE  & MAE  & $\delta _{1.05}$  & RMSE  & MAE 
 & $\delta _{1.05}$  \\ 
\midrule
CUNet         & & 3.30  & 1.66  & 97.07  & 4.90  & 2.59  & 93.74  \\
FDKN     & \multirow{8}{*}{\begin{sideways}{$\times 4 \rightarrow  \times 8$}\end{sideways}}  & 3.39  & 1.22  & 96.97  & 4.94  & 1.56  & 94.03  \\
DKN      & & 3.31  & 1.22  & 96.97  & 4.86  & 1.52  & 94.21  \\  
FDSR         & & 3.29  & 1.19  & 97.01  & 4.81  & 1.54  & 94.18  \\
DCTNet    & & 3.33  & 1.20  & 97.03  & 4.92  & 1.55 & 94.14  \\  
SUFT             &    & 3.24 & 1.20 & 97.04  & 4.79 & 1.64 & 94.08    \\
DORNet         &  & 3.22 & 1.18 & \second{97.13}  & 4.71 & 1.52 & 94.23 \\
DuCos                    & & \second{3.19}  & \second{1.17}  & 97.11  & \second{4.64}  & \second{1.46}  & \second{94.28}  \\ 
\textbf{DegBins} & & \best{3.15} & \best{1.14} & \best{97.18} & \best{4.60} & \best{1.42} & \best{94.33} \\
\midrule
CUNet        & & 5.25  & 2.69  & 92.89  & 7.79  & 3.85  & 87.19  \\
FDKN     & \multirow{8}{*}{\begin{sideways}{$\times 4 \rightarrow  \times 16$}\end{sideways}} & 5.16  & 2.34  & 92.97  & 7.56  & 2.93  & 87.91  \\
DKN      & & 5.15  & 2.32  & 93.00  & 7.54  & 2.92  & 87.93  \\  
FDSR         & & 5.16  & 2.33  & 93.01  & 7.55  & 2.92  & 87.93  \\
DCTNet    & & 5.15  & 2.32  & 92.99  & 7.55  & 2.91  & 88.02  \\  
SUFT        &      & 5.14 & 2.32 & 93.01  & 7.55 & 2.98 & 87.96                \\ 
DORNet      &   & 5.14 & 2.32 & 93.02  & 7.53 & 2.91 & 87.99              \\ 
DuCos                    & & \second{5.13}  & \second{2.31}  & \second{93.03}  & \second{7.51}  & \second{2.87}  & \second{88.08} \\ 
\textbf{DegBins} & & \best{5.06} & \best{2.28} & \best{93.12} & \best{7.47} & \best{2.85} & \best{88.14} \\
\bottomrule
\end{tabular}}
\vspace{-8pt}
\end{wraptable}
\noindent\textbf{Cross-scale DSR.} 
To validate the generalization of our method under large scale factor discrepancies, Tab.~\ref{tab_cross_scale_supp} further presents cross-scale quantitative comparisons, where the model weights trained on the $\times 4$ Hypersim dataset are directly applied to evaluate $\times 8$ and $\times 16$ DSR tasks without any fine-tuning. Compared with existing methods, our approach exhibits satisfactory cross-scale robustness. For example, DegBins achieves an average RMSE reduction of $0.17cm$ over DuCos, DORNet, SUFT, and DCTNet on the synthetic RGB-D-D dataset ($\times 4$ $\rightarrow$ $\times 8$). These results strongly demonstrate the effectiveness and superiority of our DegBins in handling scenes with large scale variations.

\begin{table}[t]
\caption{Results on real-world DSR datasets: RGB-D-D, TOFDSR, and their noisy patterns.}\label{tab_real}
\centering
\renewcommand\arraystretch{0.7}
\resizebox{0.95\linewidth}{!}{
\begin{tabular}{l|cccccccccccc}
\toprule 
\multirow{2}{*}{Method}   &\multicolumn{3}{c}{RGB-D-D}   &\multicolumn{3}{c}{TOFDSR}  &\multicolumn{3}{c}{RGB-D-D  w/ noise}   &\multicolumn{3}{c}{TOFDSR w/ noise} \\
\cmidrule(lr){2-4}\cmidrule(lr){5-7} \cmidrule(lr){8-10} \cmidrule(lr){11-13}
 &RMSE &MAE &$\delta _{1} $     &RMSE &MAE &$\delta _{1} $ 
 &RMSE &MAE &$\delta _{1} $     &RMSE &MAE &$\delta _{1} $\\ \midrule

DJF                    &5.54 &3.43 &93.81                         &5.84 &2.13 &96.79                  &7.94 &5.16 &84.82              &11.45 &7.87 &67.95     \\  
DJFR                  &5.52 &3.51 &93.58                         &5.72 &2.10 &97.03                  &7.50 &4.83 &86.25              &10.92 &7.39 &70.46     \\
CUNet                &5.84 &3.06 &94.75                         &6.04 &2.21 &96.46                  &6.69 &4.14 &89.36              &9.76 &5.86 &80.43           \\  
FDKN            &5.37 &2.70 &96.05                         &5.77 &2.19 &97.33                  &6.66 &4.26 &90.09              &8.13 &4.66 &86.24           \\
DKN             &5.08 &2.58 &96.28                         &5.50 &2.07 &97.54                  &6.50 &4.16 &90.04              &7.42 &4.29 &88.20           \\
FDSR                &5.49 &3.10 &94.77                         &5.03 &1.67 &97.61                  &6.39 &4.07 &90.69             &6.31 &3.17 &92.74           \\
DCTNet           &5.43 &3.29 &93.15                         &5.16 &2.10 &96.37                  &6.04 &3.79 &90.90             &7.52 &4.50 &86.04           \\
SFG             &3.88 &1.96 &97.09                         &4.52 &1.72 &97.45                  &5.87 &3.79 &89.84             &5.46 &2.89 &93.02           \\
DuCos           &\second{3.68} &\second{1.54} &\second{97.87}  &\second{4.29} &\second{1.15} &\second{98.67}  &\second{4.14} &\second{2.00} &\second{96.76}  &\second{5.20} &\second{2.20} &\second{96.44} \\  
\textbf{DegBins} & \best{3.52} & \best{1.45} & \best{98.01} & \best{4.06} & \best{1.03} & \best{98.87} & \best{3.90} & \best{1.91} & \best{96.94} & \best{5.04} & \best{2.11} & \best{96.59} \\
\bottomrule
\end{tabular}}
\vspace{-8pt}
\end{table}

\begin{figure}[t]
 \centering
 \includegraphics[width=0.9\columnwidth]{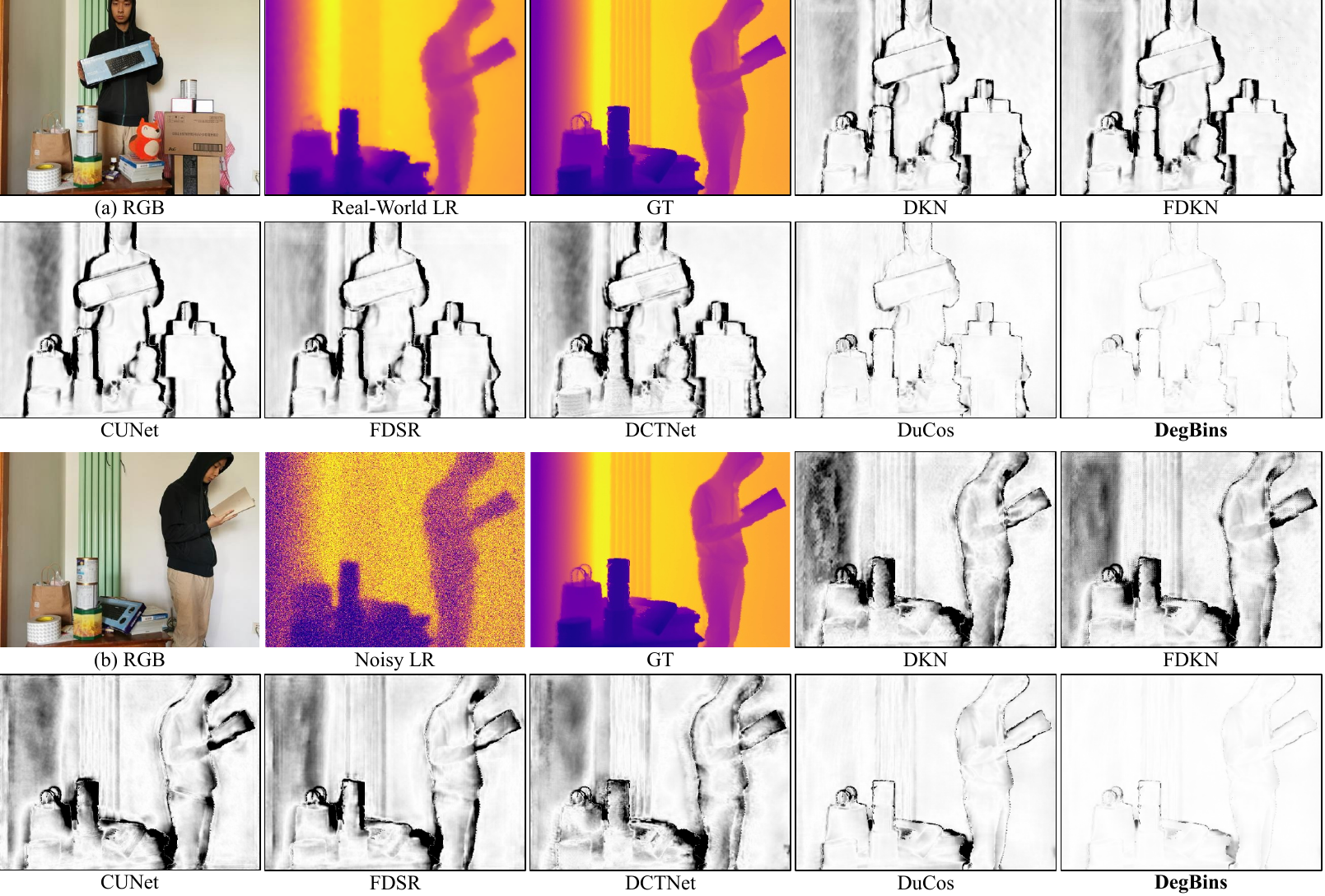}\\
 \vspace{-5pt}
 \caption{Error map comparisons on the real-world RGB-D-D dataset.}\label{fig_vis_real_rgbdd}
 \vspace{-13pt}
\end{figure}

\noindent\textbf{Real-world DSR.} 
Unlike synthetic data, real-world LR depth often suffers from unknown and complex degradations, such as noise, holes, and structural distortions, posing significant challenges for real-world DSR applications.  The first two columns of Tab.~\ref{tab_real} report the results on the RGB-D-D and TOFDSR datasets, where our method consistently outperforms all competing approaches. On TOFDSR dataset, our DegBins reduces RMSE by $0.23cm$ and MAE by $0.12cm$ compared to the suboptimal DuCos. Furthermore, the last two columns of Tab.~\ref{tab_real} present the results of joint denoising and DSR, where Gaussian blur (standard deviation $3.6$) and Gaussian noise (mean $0$ and standard deviation $0.07$) are added to the LR depth as new LR inputs. These experimental results highlight the strong noise robustness of our DegBins. For instance, compared to the second-best method, our DegBins reduces RMSE by $0.24cm$ on the RGB-D-D dataset and $0.16cm$ on the TOFDSR dataset, respectively. Fig.~\ref{fig_vis_real_rgbdd} visualizes the prediction results, further confirming that DegBins can recover sharp and accurate depth under real-world degradations.

\begin{wraptable}{r}{0.66\textwidth}
\vspace{-15pt}
\caption{Results of compressed DSR on three benchmark datasets.
}\label{tab_compressed}
\vspace{-7pt}
\Large
\centering
\renewcommand\arraystretch{0.7}
\resizebox{1\linewidth}{!}{
\begin{tabular}{l|ccccccccc}
\toprule 
\multirow{2}{*}{Method}   
&\multicolumn{3}{c}{NYU v2}   
&\multicolumn{3}{c}{RGB-D-D}  
&\multicolumn{3}{c}{TOFDSR} \\
\cmidrule(lr){2-4}\cmidrule(lr){5-7}\cmidrule(lr){8-10}
&RMSE &MAE &$\delta _{1}$      
&RMSE &MAE &$\delta _{1}$  
&RMSE &MAE &$\delta _{1}$    
\\ \midrule
DJF     &119.41 &101.14 &8.53  &34.48 &29.93 &16.26 &47.98 &44.30 &10.94 \\
DJFR    &147.43 &121.73 &7.85  &40.46 &34.49 &14.03 &50.14 &45.81 &8.59  \\
CUNet   &133.90 &108.96 &7.14  &35.03 &30.67 &15.08 &50.25 &46.53 &8.96  \\
FDKN    &187.53 &154.15 &4.26  &42.37 &36.91 &14.04 &49.34 &44.88 &14.28 \\
DKN     &186.10 &152.74 &4.33  &40.38 &35.15 &14.40 &47.27 &43.34 &9.58  \\  
FDSR    &118.76 &101.28 &8.39  &35.88 &30.15 &20.64 &45.96 &40.92 &20.26 \\  
DCTNet  &118.45 &101.30 &8.20  &37.85 &33.25 &12.82 &51.58 &48.41 &6.88  \\  
DORNet  &128.25 &106.11 &8.58  &41.13 &35.07 &15.28 &53.77 &47.22 &12.92 \\ 
DuCos   &\second{58.74} &\second{46.00} &\second{26.19} 
        &\second{24.64} &\second{21.13} &\second{26.45} 
        &\second{30.40} &\second{25.63} &\second{22.37} \\ 
\textbf{DegBins} 
        &\best{51.26} &\best{43.48} &\best{28.81} 
        &\best{19.74} &\best{18.41} &\best{30.63} 
        &\best{27.35} &\best{24.52} &\best{22.80} \\
\bottomrule
\end{tabular}}
\vspace{-5pt}
\end{wraptable}
\noindent\textbf{Compressed DSR.} 
Compression is often necessary for efficient depth transmission, particularly in real-time downstream applications. However, the compression inevitably introduces artifacts and degrades depth quality, making accurate HR depth reconstruction from compressed LR depth crucial for real-world deployment. Following~\cite{zheng2025decoupling, yan2025ducos}, we evaluate DegBins on the compressed DSR task. Tab.~\ref{tab_compressed} shows that DegBins consistently achieves competitive performance on all datasets, surpassing the suboptimal method by $7.48cm$, $4.90cm$, and $3.05cm$ in RMSE on NYU v2, RGB-D-D, and TOFDSR, respectively. These results validate the superiority of DegBins in compressed scenarios.

\subsection{Ablation Study}
The baseline removes all bin-related modules from DegBins, retaining only the gray branch in Fig.~\ref{fig_pipeline}.

\begin{figure}[t]
\centering
\begin{minipage}[t]{0.491\linewidth}
    \centering
        \includegraphics[width=1\linewidth]{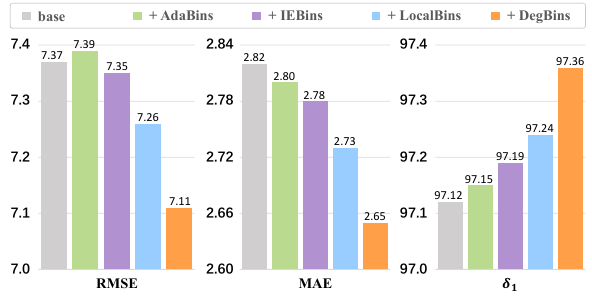}
\caption{Ablation of different bin strategies on the synthetic NYU v2 ($\times 16$) dataset.}\label{fig_ab_different_bins}
\end{minipage}
\hspace{1.5pt}
\begin{minipage}[t]{0.491\linewidth}
    \centering
    \includegraphics[width=1\linewidth]{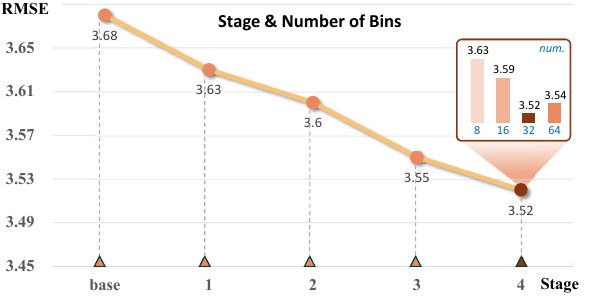}
    \caption{Ablation of the stage and number of bins on the real-world RGB-D-D dataset.}\label{fig_ab_bin_stage_num}
\end{minipage}
\vspace{-6pt}
\end{figure}

\noindent \textbf{Bin Types.} 
Fig.~\ref{fig_ab_different_bins} compares our DegBins with previous bin-based methods, including AdaBins~\cite{bhat2021adabins}, IEBins~\cite{shao2023iebins}, and LocalBins~\cite{bhat2022localbins}. To ensure a fair comparison, we directly replace our DegBins strategy with each of these methods while keeping the rest of the network architecture unchanged. These experimental results demonstrate that our DegBins achieves the best performance across all evaluation metrics. On average, our method outperforms the competing approaches by $0.23cm$ in RMSE, $0.12cm$ in MAE, and $0.17$ percentage points in $\delta _{1} $. These results confirm that our approach effectively reduces DSR error and improves reconstruction quality.

\noindent \textbf{DDB Details.} 
As illustrated in the line chart of Fig.~\ref{fig_ab_bin_stage_num}, we conduct an ablation analysis with different numbers of stages. Here, `2-Stage' indicates that our method only employs the `Layers' from the first two stages along with their corresponding DDBs (see Fig.~\ref{fig_pipeline}).  It can be clearly observed that as the number of stages increases, the RMSE consistently decreases. When all four stages are introduced, our DegBins achieves the best performance, surpassing the baseline by $0.16cm$ RMSE on the real-world RGB-D-D dataset. Therefore, we adopt four stages as the default setting. 
In addition, the bar chart in Fig.~\ref{fig_ab_bin_stage_num} presents a quantitative comparison for different numbers of bins.  These results show that DSR performance gradually improves as the number of bins increases, while it begins to decline once the number exceeds $32$.  As a result, we adopt $32$ bins for DegBins.

Furthermore, Tab.~\ref{tab_ab_deg} reports the impact of degradation representations on bin construction. Compared with Baseline (i), the results show that introducing degradation representations enables adaptive modulation of both bin ranges (ii) and probability distributions (iii), significantly improving depth reconstruction accuracy. When modulating both components simultaneously, our method (iv) achieves optimal performance, outperforming the baseline by $0.26cm$ in RMSE on the $\times16$ NYU v2 dataset and by $0.16cm$ in RMSE on the real-world RGB-D-D dataset.

\begin{table}[t]
\caption{Ablation of the impact of degradation on synthetic NYU v2 and real-world RGB-D-D.
}\label{tab_ab_deg}
\centering
\small
\renewcommand\arraystretch{0.75}
\resizebox{0.95\linewidth}{!}{
\begin{tabular}{l|cc|cccccc}
\toprule 
\multirow{2}{*}{Method}   
& \multirow{2}{*}{Deg. in Bins}
& \multirow{2}{*}{Deg. in Prob. Dist.}
&\multicolumn{3}{c}{NYU v2}   
&\multicolumn{3}{c}{RGB-D-D}  \\
\cmidrule(lr){4-6}\cmidrule(lr){7-9}
&&& RMSE & MAE & $\delta _{1}$ & RMSE & MAE & $\delta _{1}$
\\ \midrule
i && & 7.37  & 2.82  & 97.12 & 3.68  & 1.54  & 97.87  \\
ii & \checkmark &  & 7.28  & 2.76  & 97.19 & 3.63  & 1.50  & 97.94 \\
iii &  & \checkmark  & 7.20  & 2.74  & 97.25 & 3.59  & 1.48  & 97.96 \\
iv & \checkmark  & \checkmark & 7.11  & 2.65  & 97.36 & 3.52  & 1.45  & 98.01 \\
\bottomrule
\end{tabular}}
\vspace{-6pt}
\end{table}

\section{Conclusion}
\label{sec_conclusion}
In this paper, we present DegBins, a novel and generalizable DSR framework based on residual modeling. By reformulating DSR as a hybrid regression-classification task, DegBins facilitates more flexible and expressive residual representations. Furthermore, we introduce a multi-stage degradation-driven binning strategy that leverages high-dimensional local degradation cues to adaptively adjust bin ranges and optimize their probability distributions. Extensive experiments validate the superiority of DegBins in terms of accuracy, robustness, and generalization ability. 

\noindent \textbf{Limitation.} 
Although DegBins achieves strong performance, it incurs relatively high computational complexity (see Appendix~\ref{appendix_comp}) due to the heavy feature extraction backbone. In future work, we plan to explore lightweight feature extraction architectures to enable real-time performance. 

\noindent \textbf{Broader Impact.} 
This work improves the accuracy, robustness, and generalization of depth super-resolution, benefiting many downstream applications such as robotics, augmented reality, and 3D scene understanding, where precise geometric perception is essential.


{   
\bibliographystyle{ieeenat_fullname}
\bibliography{ref}
}


\newpage
\appendix

\section{Technical Appendices for DegBins}
\label{appendix}

\subsection{Metrics} 

Given the predicted depth map $\mathbf{X}$ and the corresponding GT $\mathbf{Z}$ containing $m$ valid pixels, we evaluate performance using RMSE (cm), MAE (cm), and the accuracy metric $\delta_1$, defined as follows:
\begin{equation}
    \begin{split}
    &\text{MAE}: \frac{1}{m}\sum\limits{{{\left| \mathbf Z-{\mathbf X} \right|}}},  \\
    &\text{RMSE}: \sqrt{\frac{1}{m}\sum\limits{{{\left( \mathbf Z-{\mathbf X} \right)}}}^2}, \\
    &{\delta }_{i}: \frac{r}{m}\times 100\%, \ r: \max \left( {{\mathbf Z}/{\mathbf X},{\mathbf X}/{\mathbf Z}} \right)<{1.05}^{i}.
    \end{split}
\end{equation}
Since the values of ${\delta }_{2}$ and ${\delta }_{3}$ are typically very close across different DSR methods, we primarily report ${\delta }_{1}$ for comparison.

\subsection{Implementation Details}
\label{Sec:Implementation}
DegBins is implemented in PyTorch and trained on four NVIDIA RTX A6000 GPUs. The model is optimized for 200 epochs using the Adam optimizer~\cite{Kingma2014Adam}, with an initial learning rate of $5 \times 10^{-5}$, which is reduced by a factor of 2 at epochs 40, 80, and 120. For data augmentation, we adopt random horizontal flipping and random $90^\circ$ rotations to improve robustness. Since Hypersim~\cite{roberts2021hypersim} provides $1024 \times 768$ high-resolution images, while the test sets of other benchmarks are typically of much lower resolution, we randomly crop $256 \times 256$ patches during training on Hypersim to ensure efficiency and improve generalization.

\subsection{Complexity Analysis}\label{appendix_comp}
\begin{wraptable}{r}{0.65\textwidth}
\vspace{-12pt}
\caption{Complexity comparisons on the real-world RGB-D-D dataset. All methods are measured using a single 4090 GPU.
\vspace{-5pt}
}\label{tab_complexity_appendix}
\small
\centering
\renewcommand\arraystretch{1.1}
\resizebox{1\linewidth}{!}{
\begin{tabular}{l|cccc}
\toprule 
\multirow{2}{*}{Method}  & Parameters  & Time  & Speed  & RMSE  \\ 
\cmidrule(lr){2-5}
 & (M)  & (ms)  & (FPS)  & (cm)  \\\midrule
DCTNet \cite{zhao2022discrete}    & \best{0.48}  & 9.15   & 109.29  & 5.43  \\
FDKN \cite{kim2021deformable}     & 0.69  & \second{5.78}   & \second{173.01}  & 5.37  \\
DKN \cite{kim2021deformable}      & 1.16  & 17.75  & 56.34   & 5.08  \\
FDSR \cite{he2021towards}         & \second{0.60}  & \best{5.05}   & \best{198.02}  & 5.49  \\
SUFT \cite{shi2022symmetric}      & 22.01 & 13.33  & 75.19   & 5.41 \\
SGNet \cite{wang2024sgnet}        & 8.97  & 33.94  & 29.46   & 5.32  \\
SFG \cite{yuan2023structure}      & 63.55 & 21.81  & 45.85   & 3.88  \\
DuCos~\cite{yan2025ducos}    & 34.38  & 25.09  & 39.86  & 3.68 \\
\midrule
\textbf{DegBins} (RG~\cite{zhang2018image}) & 16.28  & 9.75  & 102.6  & \second{3.63} \\
\textbf{DegBins} (DuCos)  & 42.15  & 34.64  & 28.87  & \best{3.52} \\
\bottomrule
\end{tabular}}
\vspace{-5pt}
\end{wraptable}
Tab.~\ref{tab_complexity_appendix} presents a detailed complexity comparison on the real-world RGB-D-D dataset. DegBins first adopts DuCos~\cite{yan2025ducos} as the feature extraction backbone in Fig.~\ref{fig_pipeline}, which incorporates the large-scale Depth Anything v2~\cite{yang2025depth}. While DegBins achieves the lowest error, it incurs higher computational cost and slower inference speed due to the substantial size of the depth foundation model. 
As an alternative, we replace the backbone with a lightweight residual group~\cite{zhang2018image} for feature extraction. With only a marginal drop in accuracy, the number of parameters is reduced from 42.15M to 16.28M, and the inference speed improves by approximately 74 FPS. Moreover, DegBins-RG attains a lower RMSE than DuCos. These results suggest that exploring more efficient architectures with stronger accuracy–efficiency trade-offs is a promising direction for future work.


\end{document}